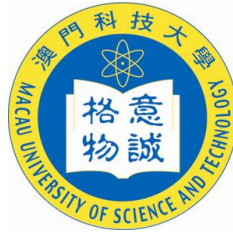

# 澳門科技大學
# MACAU UNIVERSITY OF SCIENCE AND TECHNOLOGY

Thesis for Bachelor of Science
Faculty of Information Technology

Title: **<u>Dynamic Decision Process Modeling and Relation-line Handling in Distributed Cooperative Modeling System</u>**

Name: <u>王孟函</u>
Student No: <u>0609853Q-I011-0015</u>
Supervisor: <u>Prof. Cai Zhiming (蔡智明)</u>

**Apr 2010**

# 摘要

分布式協作建模系統(DCMS)通過基於網路的分布式建模及多模板集結以解決涉及多位參與者的複雜決策問題。本文以擴展 DCMS 並使之支持動態決策過程為目標。

本文首先討論了馬爾可夫決策過程(MDP)的特點及最優策略的尋求方法，同時也簡略介紹了本質上與 MDP 等價的動態貝葉斯決策網路。接下來，本文探討並實現了基於離散和連續狀態變數的 MDP 過程中的預測以及若干種多指標間的關聯度分析方法，後者有助於決策者瞭解各指標間的相互作用關係以便於制定合理的策略。

本文亦介紹了作為 DCMS 擴展的歷史資料庫處理工作。另一項工作為對 DCMS 中的繪圖類繼承架構進行的重組，及在此基礎上實現的曲線關聯線的處理。

Key words: 分布式協作建模系統, 動態決策, 馬爾可夫決策過程, 關聯度分析, 關聯線

# Abstract

The Distributed Cooperative Modeling System (DCMS) solves complex decision problems involving a lot of participants with different viewpoints by network based distributed modeling and multi-template aggregation.

This thesis aims at extending the system with support for dynamic decision making process. First, the thesis presents a discussion of characteristics and optimal policy finding Markov Decision Process as well as a brief introduction to dynamic Bayesian decision network, which is inherently equal to MDP. After that, discussion and implementation of prediction in Markov process for both discrete and continuous random variable are given, as well as several different kinds of correlation analysis among multiple indices which could help decision-makers to realize the interaction of indices and design appropriate policy.

Appending history data of Macau industry, as the foundation of extending DCMS, is introduced. Additional works include rearrangement of graphical class hierarchy in DCMS, which in turn allows convenient implementation of curve relation-line, which makes template modeling clearer and friendlier.

Key words: DCMS (Distributed Cooperative Modeling System), Dynamic Decision, MDP(Markov Decision Process), Correlation Analysis, Relation-line

# List of Contents

# 1  Introduction

## 1.1  Distributed Cooperative Modeling System

The Distributed Cooperative Modeling System (DCMS) is developed to provide an integrated system with combination of modeling and decision in order to aid decision-makers to analyze and design policies for complex problems. The system has the following characteristics [1]:

1.  Visualized graphical modeling mechanism. It helps to construct intuitional graphical models for large and complex problems with several simple basic elements as well as different kinds of relationship among them.

2.  Network based distributed cooperation. Since analysis and decision making of large and complex problems usually calls for cooperative work of one or several teams, the distributed system facilitates communication among users.

3.  Aiding for decision making. The system manages and organizes data generated in the modeling process and provides several kinds of decision-support algorithm, which can automatically aggregate multiple models from different users to maximize the overall benefit.

4.  Management of participants' working status. The system records and organizes the working history of participants and provides statistical information to managers, helping in project management and supervision.

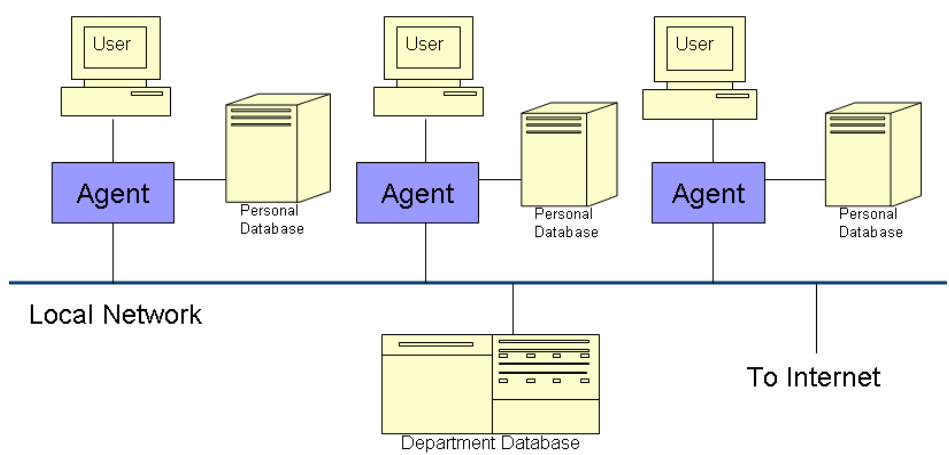

Figure 1-1 Structure of DCMS on local network

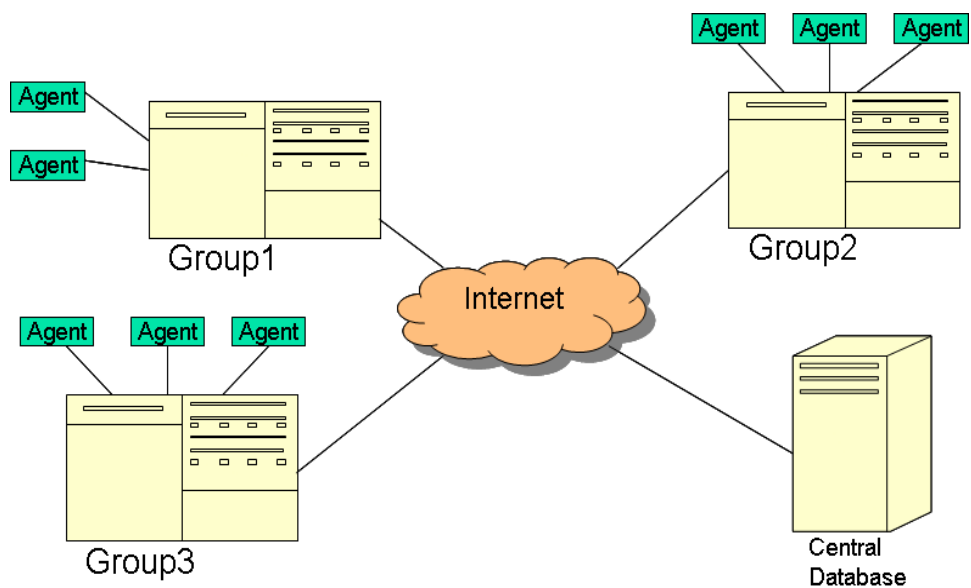

Figure 1-2 Structure of DCMS on internet

By using DCMS, decision-making agents within the same department store their modeling templates both in their local databases and the department database. The databases from different departments could then be connected with internet for knowledge sharing and synthesizing.

The template built in DCMS is the graphical representation from an agent for relationship among the elements in a certain decision problem. The template can support inferences such as optimal solution selection, reasoning on direct or indirect relationship among the elements, as well as dividing and reorganizing of the template in order to represent complex models in multiple views and levels.

The elements and relationships are the semantic primary of DCMS. I will introduce the goal, solution and condition elements as well as relationship line, which are essential in decision-making.

1. Goal

   Goal is the ultimate objective of analyzing and decision making. Different goals may give rise to conflictions. Possible goals include feasibility, reliability, economic efficiency, etc.

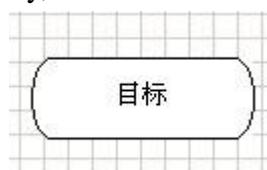

Figure 1-3 Goal

2. Solution

Solution is the possible way to achieve expected goal, the object of decision-making. The decision problem may present several candidate solutions and the decision process tries to find the optimal solution for the goal.

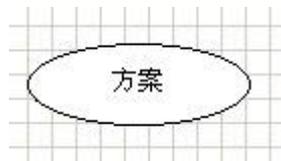

Figure 1-4 Solution

3. Condition

Condition is the objective limit on agent, goal and solutions, such as time, fund and human resource. The decision process should find a optimal solution feasible under the given condition.

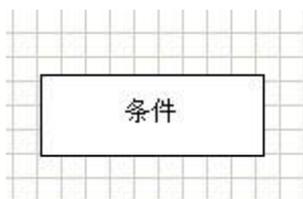

Figure 1-5 Condition

4. Relation line

Relation line denotes an element is relative to another. The relation lines are usually assigned numerical support values, which are utilized in inference.

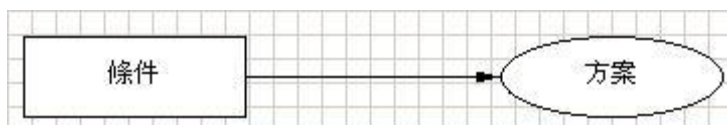

Figure 1-6 Relation line

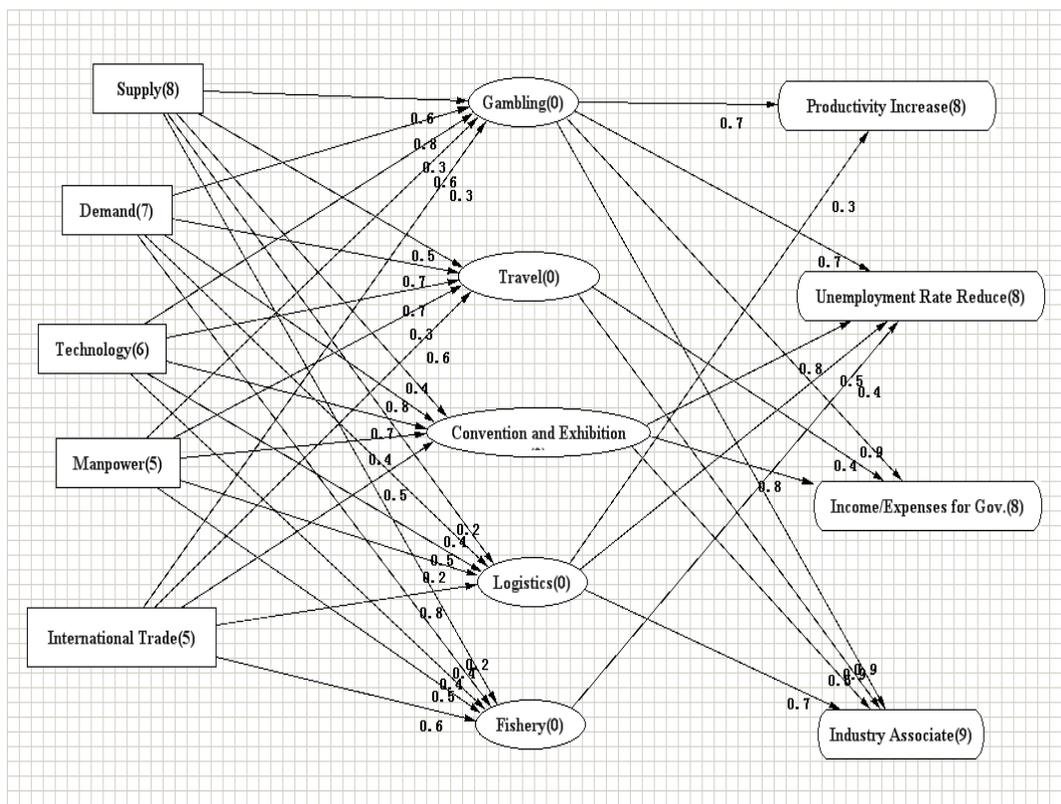

Figure 1-7 A template with goals, conditions, solutions and relation-lines

The main algorithms used in DCMS to achieve synthesized decision are Analytical Hierarchy Process (AHP) for single template and Ordered Weighted Geometric (OWG) for multiple templates.

AHP is a system analytic method which combines qualitative and quantitative analysis. It decomposes the problem into several components, groups components into a progressive level model, assigns quantitative relative significance and weight to each level and finally computes the synthesized evaluation of the lowest level -- solution level.

OWG provides synthesizing of preference of different participants. The OWG operator is based on fuzzy logic conception, which provides a flexible tool to represent the fuzziness of human thought.

## 1.2 Dynamic Decision Process

Decision-making generally refers to the human activity of choosing an action under certain objective conditions in order to optimize the achievement of an expected goal. In practice, decisions are not always made in one-shot -- there are a lot of dynamic decision problems, that is, decisions are made sequentially based on changing of environment and decision agent states in time-series.

Moreover, the decision process may not be totally deterministic: due to the complexity of such problems, people generally do not have accurate knowledge about cause and effect, and/or the cost to analyze a large amount of details does not pay. That is to say, the consequence of actions can only be measured with probability method. A specification of the outcome probabilities for each action in each possible state is called a transition model [2].

For simplicity, the Markov assumption is usually made, which assumes that the state at the next time point is only affected by the current state and action taken at current time point and not by earlier history.

In order to measure the degree to which the whole decision process contributes to our expected goal, utility, which captures the decision agent's preference between states or state sequences, is used. In order to match a specific state sequence to an appropriate utility value, a general approach is to assign a reward to each state, and define the utility as accumulated rewards. Under stationary consideration, which means two state sequences beginning with the same state should be preferred ordering in the same way as their sub-sequences without the start state, there are two ways to aggregate rewards:

1. Additive rewards:
   Utility = Reward($State_{t0}$) + Reward($State_{t1}$) + Reward($State_{t2}$) + …
2. Discounted rewards:
   Utility = Reward($State_{t0}$) + $\gamma$Reward($State_{t1}$) + $\gamma^2$ Reward($State_{t2}$) + …, where $0<\gamma<1$.

The discounted rewards are helpful at introducing a limit to the final utility and assigning more importance to nearer future.

An optimal solution to a dynamic decision problem consists of sequential decisions which yield the maximum expected utility. However, as mentioned before, a decision cannot guarantee a definite consequence, and a fixed decision sequence may not always conform to the actual situations. As a result, rather than fixing the decision as in static decision problems, a function should be defined to assign appropriate action for each possible state. This kind of solution is called a policy. With a policy, the current state is mapped to an action, which is executed, and then the consequent state is observed and mapped to the next action.

Another issue in designing the dynamic decision process is the representation of states. A state should contain values of all the relative indices which are significant enough for describing the current environment and making decisions.

One approach is aggregating all the information into a single random variable. For example, if there are N indices for the state and each of them has M possible values, then the aggregated state variable has $N^M$ possible values. Under the Markov assumption, the current state and action could be used to define the transition model and get a Markov decision process, or MDP.

On the other hand, if each individual index is represented along with conditional dependency among them, a dynamic Bayesian decision network is formed.

### 1.2.1 Markov Decision Process

An MDP can be defined with the following components:

- Initial State: $S_0$
- Transition Model: T(s, a, s')
- Reward function: R(s)

Where s and s' stands for current and next state respectively, a stands for the action and T(s, a, s') denote the joint probability distribution of s', s and a.

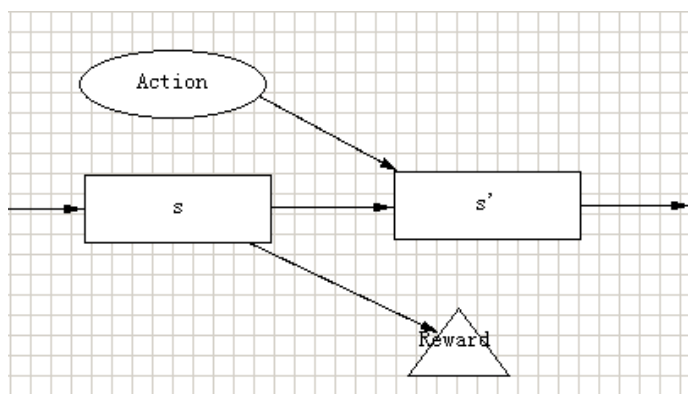

Figure 1-8 A slice of an MDP

The value of a policy is the expected sum of rewards obtained over all possible consequent state sequences, using the discounted rewards. Here we introduce a widely used algorithm, value iteration, for finding an optimal policy.

Value iteration computes the utility for each state. It should be noticed that utility of a state is totally different from reward of a state: it is the expected sum of rewards computed on all the possible following sequences depending on the policy used. If we know the utility of each state, we only need to select the action for each state that maximizes the expected utility of the next state.

The utility of a state clearly equals to reward of that state plus the maximum expected utility from the next state, which gives us the following representation for state s:

$$U(s) = R(s) + \gamma \max_a \sum_{s'} T(s, a, s') U(s') \qquad (1.1)$$

Which is called the Bellman equation [*]. The utilities of the states under the optimal policy are the unique solutions of all these Bellman equations.

In order to solve these equations, we initialize the utilities to arbitrary values, and replace the left-hand side by the computed-result from right-hand side. This iteration step is called Bellman Update. After a certain number of iterations equilibrium will be reached, where the final utility values are the solutions. Then the optimal policy $\pi^*$ can be found using the following equation:

$$\pi^*(s) = \text{argmax}_a \sum_{s'} T(s, a, s') U(s') \qquad (1.2)$$

### 1.2.2 Dynamic Bayesian Decision Networks

A dynamic Bayesian decision network is a Bayesian network representing sequential probability model along with action and reward nodes.

In fact, every dynamic Bayesian network can be converted to a Markov chain by combining all the indices into a single state. The advantage of using dynamic Bayesian network is that the conditional dependencies among the indices can greatly reduce the model inference space as well as express the causal relationship clearly [3]. Nevertheless, to construct the structure of the dynamic Bayesian network is complicate and usually requires assistance of domain experts. Here we will give a brief introduction to dynamic Bayesian decision network along with some background concepts.

By the noun "probability" we usually mean prior probability, the degree of belief about a random variable in the absence of any other information. On the other hand, Conditional probability is related with value of other random variables. For example, the conditional probability P(a|b) indicates the probability of a given that all we know is b. Conditional probabilities can be defined in terms of prior probabilities in the form P(a|b) × P(b) = P(a, b). It comes from the fact that, for a and b to both be true, b need to be true, and a should to be true given b. A simple inference will yield Bayes' theorem:

$$P(H|X) = P(X|H)\ P(H)\ /\ P(X) \qquad (1.3)$$

It provides us a way of calculating the posterior probability $P(H|X)$ from $P(H)$, $P(X)$ and $P(X|H)$ [4].

Two variables are conditional independent if they are independent given the value of a third variable. Independence and conditional independence relationships among variables can greatly reduce the number of probabilities that need to be specified.

A Bayesian network is a directed acyclic graph with conditional probability tables (CPT) for each node. The random variables make up the nodes of the network, while a set of directed links connects pairs of nodes denoting direct influence. Each node X has a conditional probability distribution $P(X|Parents(X))$ that quantifies the effect of the parents on the node. Each row in a CPT contains the conditional probability of each node value for a possible combination of parent values. Suppose the world is composed of random variables $Y_1$, ..., $Y_n$, the probability $P(Y_1=y_1, ..., Y_n=y_n)$ can be given by the formula

$$P(y_1, ..., y_n) \quad = \prod_{i=1}^{n} P(y_i|parents(Y_i)) \qquad (1.4)$$

Thus, every probability value that can be represented in the joint distribution can also represented by the product of appropriate elements of CPTs.

Under Markov assumption, the nodes and links are identical in each time slice's Bayesian network. Adding parent links to state indices in the immediate preceding time slice, we can easily convert a static Bayesian network to a dynamic one. After adding the action and reward node, similar methods as in MDP could be used to infer, predict and make decisions in dynamic Bayesian decision network.

## 1.3   Dynamic Decision in Macau Industry Diversification

The current mainstay industry in Macau is undoubtedly the gambling industry. The unitary industry structure as well as potential unstableness of gambling makes the development of economy in Macau unsustainable. As a result, Macau industry diversification is proposed as a solution.

Previous researches on the problem of Macau industry diversification were mostly done through traditional observation and analysis. Nevertheless, as the advancement of economy and technology as well as social environment, the factors affecting industry structure are changing all the time, and policy for industry should be adjusted accordingly. Therefore, the dynamic decision approach could be highly useful.

As previously presented, DCMS already has functions as visual modeling and static decision optimization, as well as synthesizing of opinions from multiple participants. Under Markov assumption, the dynamic decision process could be easily represented as a template by modeling any single slice of the process along with transition model. In a word, DCMS offers a good framework for extension of dynamic decision implementation and its application.

# 2 Prediction in MDP

Without considering the action and reward nodes, the dynamic decision process will become a problem of objective probabilistic reasoning over time. One of the most useful inference mechanisms is prediction: computing the probability distribution of a future state based on history data. In the practical problem of Macau industry diversification, by predicting the possible value of importance industrial indices supposing no actions are taken, we can realize the existing problem in current industry structure and think up corresponding solutions.

In MDP, if $\mathbf{P}(\mathbf{X}_{t+1}|\mathbf{x}_t)$ denotes the probability distribution of state at time t+1 defined for each value in $\mathbf{x}_t$, which is the collection of possible state values at time t, the predicted distribution at time k starting from state $x_0$ will be

$$P(\mathbf{X}_k) = \sum_{t=0}^{k} \mathbf{P}(\mathbf{X}_{t+1}|\mathbf{x}_t) \tag{2.1}$$

Nevertheless, if the state variable is continuous rather than discrete, the summation on the right-hand side should be replaced by an integral. Next, prediction for discrete variable and continuous variable will be respectively introduced.

## 2.1 Prediction from History Data for Discrete State Variable

Handling the prediction for discrete state variable can actually be quite elegant: the summation mentioned above can be turned into arithmetic of matrices. We use **P** to denote the state transition probability matrix. **P** consists of n rows and n columns, where n is equal to the number of possible states. A cell of **P**, $p_{ij}$, is $P(X_{t+1} = j \mid X_t = i)$, that is, the probability of transition from state i to state j. The formula for getting **P** from the history data is:

$$p_{ij} = P(X_{t+1} = j | X_t = i) = \sum_t \hat{N}(X_{t+1} = j, X_t = i) / \sum_t \hat{N}(X_t = i) \quad (2.2)$$

where $\hat{N}(x)$ stands for the number of appearances that x made in the history data. Therefore we can see the learning process for discrete MDP is simply counting. One thing to be noticed is that if a certain state did not appear in the history data, its relative cells will be zero and the Markov process will never transit to that state using the computed transition matrix. If the training data is large enough, we can plus one to each count to avoid this problem.

After obtaining the transition matrix, the prediction process is quite straight-forward. The probability distribution of time t+1 is the probability distribution of time t multiplied by the transition matrix. Consequently, the probability distribution of time t+k is

$$\mathbf{X}_{t+k} = \mathbf{X}_t \, \mathbf{P}^k \qquad (2.3)$$

Some may notice that our start point of inference, which is the last state in history data, is a definite value rather than a distribution vector. This equals to a probability of 1 for that value, and 0 for all other values. For instance, if there are 3 states A, B, C, and the start state of prediction is C, we should represent it as vector (0, 0, 1).

Still, there are some problems to settle when applying this prediction process to our application, Macau industry diversification. Industrial indices are by nature continuous. Only by discretizing them into discrete values can we apply the presented matrix-based approach. Another issue is that the industrial situation almost always involves multiple indices. Therefore a method must be designed to synthesize these indices into a single state variable. One feasible approach is as following:

1. Divide each continuous index into several levels and map continuous history values to discrete levels with fuzzy membership function. Thus we get fuzzy level membership vector for each index; by listing them together and treating them as rows of a matrix, we get fuzzy membership matrix for all the indices.

2. Assign weight to each index and multiply the membership matrix by the weight vector. Then we get the synthesized level membership vector.

3. Use maximum membership principle to assign a unique level to each history time entry.

After these three steps, the history data is perfectly discrete and it is easy to compute the transition matrix, where the number of state is the number of levels. However, this process often calls for human interception. The leveling of indices should be done by the system user, and this leveling result is of great importance to the whole prediction process. In a word, the performance of prediction, to a large extent, depends on the user's leveling action.

## 2.2 Prediction from History Data for Continuous State Variable

How to define the transition model for continuous state variable? Clearly the transition matrix could no longer be used, since the value of continuous variable cannot be iterated. In fact, an approximate distribution for the state variable and an appropriate function form to model relationship between neighboring states should be chosen.

For simplicity and objectivity, in this thesis linear Gaussian model is used to analyze a single state variable to demonstrate learning and prediction for continuous state variable. For more than one index, joint distribution to substitute single variable distribution could be used. The linear Gaussian model could also be substituted by other function and distribution models and the prediction process will be similar.

Suppose $x_{t+1}$ is a linear function of $x_t$ affected by a Gaussian noise. The transition model is

$$P(x_{t+1}|x_t) = N(\mu, \sigma^2)(x_{t+1}) = \frac{1}{\sigma\sqrt{2\pi}}e^{-\frac{1}{2}\left(\frac{x_{t+1}-\mu}{\sigma}\right)^2}, \ \mu = ax_t + b \quad (2.4)$$

There are three unknown parameters: a, b and σ. In order to find values for these parameters, learning is to be performed on history data by using the following method.

We use **d** to describe the history data -- all the $(x_{t+1}, x_t)$ pairs in history, $h_{a,b,\sigma}$ to describe the heuristic of a, b, σ values. According formula (1.3), since P(**d**)=1,

$$P\left(h_{a,b,\sigma}\big|d\right)=P\left(d\big|h_{a,b,\sigma}\right)P\left(h_{a,b,\sigma}\right) \tag{2.5}$$

The most likely heuristic of a, b, σ has the maximum probability $P\left(h_{a,b,\sigma}\big|\mathbf{d}\right)$. On the right-hand side, it is relatively easy to compute $P\left(\mathbf{d}\big|h_{a,b,\sigma}\right)$, but the prior probability $P\left(h_{a,b,\sigma}\right)$ is rather difficult to find, the maximum-likelihood assumption, that is, all the heuristics have equal likelihood, is assumed. Therefore, the only term that needs to maximized is

$$P\left(d\big|h_{a,b,\sigma}\right)=\prod_{t=0}^{n-1}\frac{1}{\sigma\sqrt{2\pi}}e^{-\frac{1}{2}\left(\frac{x_{t+1}-(ax_t+b)}{\sigma}\right)^2} \tag{2.6}$$

Where n is the number of history data records.

Since it is hard to maximize the product value, the product could be converted to a sum − log likelihood and maximizing it is equivalent to maximize the original probability.

$$L=\sum_{t=0}^{n-1}\log\frac{1}{\sigma\sqrt{2\pi}}e^{-\frac{1}{2}\left(\frac{x_{t+1}-(ax_t+b)}{\sigma}\right)^2}=N\left(-\log\sqrt{2\pi}-\log\sigma\right)-\sum_{t=0}^{n-1}\frac{[x_{t+1}-(ax_t+b)]^2}{2\sigma^2} \tag{2.7}$$

The a, b, σ values that maximize L are the solutions of the equation set got by compute the log likelihood's partial derivatives with respect to each of three parameters and assign them the value 0:

$$\frac{\partial L}{\partial a}=-\frac{1}{\sigma^2}\sum_{t=0}^{n-1}(ax_t^2+bx_t-x_tx_{t+1})=0$$

$$\frac{\partial L}{\partial b}=-\frac{1}{\sigma^2}\sum_{t=0}^{n-1}(ax_t+b-x_{t+1})=0$$

$$\frac{\partial L}{\partial \sigma}=-\frac{n}{\sigma}+\frac{1}{\sigma^3}\sum_{t=0}^{n-1}[x_{t+1}-(ax_t+b)]^2=0 \tag{2.8}$$

The solutions for (2.8) are

$$a=(n\sum_t x_t x_{t+1}-\sum_t x_t\sum_t x_{t+1})/[n\sum_t x_t^2-(\sum_t x_t)^2]$$

$$b=(\sum_t x_{t+1}-a\sum_t x_t)/n$$

$$\sigma = \sqrt{\sum_t [x_{t+1} - (ax_t + b)]^2 / n} \tag{2.9}$$

To do prediction, as mentioned before, integral should be used to compute distribution of next time slice from distribution current time slice and transition model:

$$P(x_{t+1}) = \int_{x_t} P(x_{t+1}|x_t) P(x_t) dx_t \tag{2.10}$$

If $x_t$ is a definite value, for example, the last record in the history data, $P(x_t) = 1$ and $P(x_{t+1}) = P(x_{t+1}|x_t)$. If $x_t$ is a Gaussian distribution with expectation $\mu_t$ and standard deviation $\sigma_t$, the integral value will turn out to be

$$P(x_{t+1}) = \alpha e^{-\frac{1}{2}\left(\frac{x_{t+1} - (a\mu_t + b)}{a^2 \sigma_t^2 + \sigma^2}\right)^2} \tag{2.11}$$

From which it can concluded that the new state is also a Gaussian distribution with a new expectation linear to $\mu_t$ and an updated standard deviation.

## 2.3 Demonstration

The Implementation uses linear Gaussian to give prediction distribution on continuous variable based on history data for future five years.

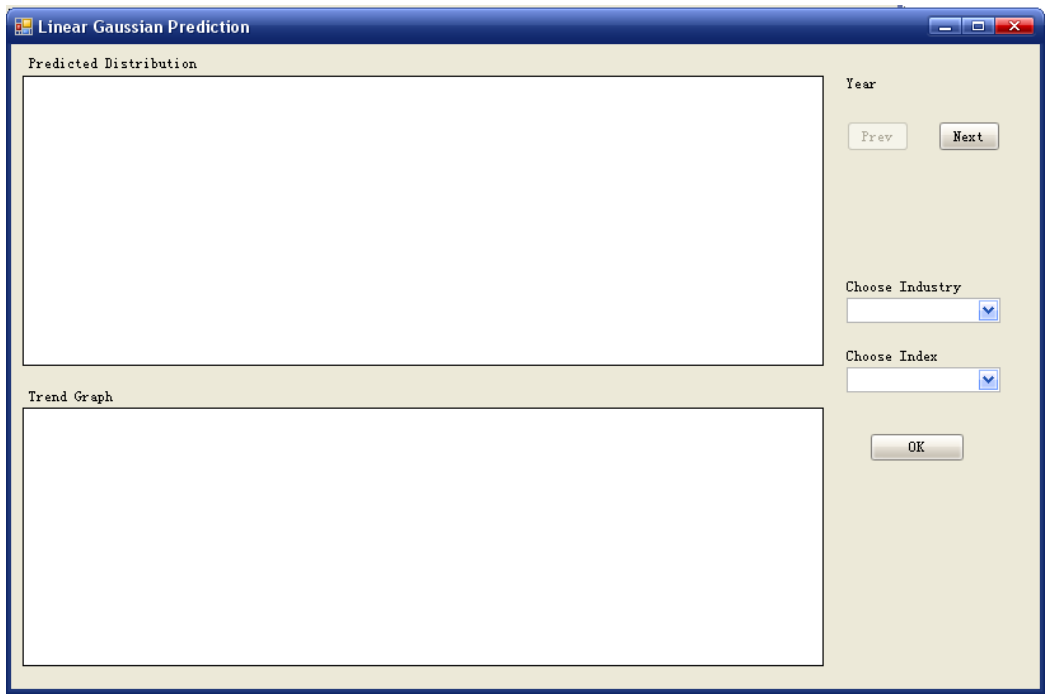

Figure 2-1 Prediction interface

Figure 2-1 is the prediction interface. The left part is used to plot prediction result. The panel on the top shows predicted distribution, and the one on the bottom shows the trend graph of the index. At the top right corner, users can see and control year of the shown distribution is. From the two combo-boxes users can choose which index of which industry to predict.

The following is a demonstration of prediction of employed population of Macau manufacturing industry. The history data is from 2002 to 2008.

| Year | 2002 | 2003 | 2004 | 2005 | 2006 | 2007 | 2008 |
|---|---|---|---|---|---|---|---|
| Employed Population (thousand) | 42.0 | 37.7 | 36.1 | 35.3 | 29.5 | 24.0 | 24.6 |

Table 2-1 History data of manufacturing employed population

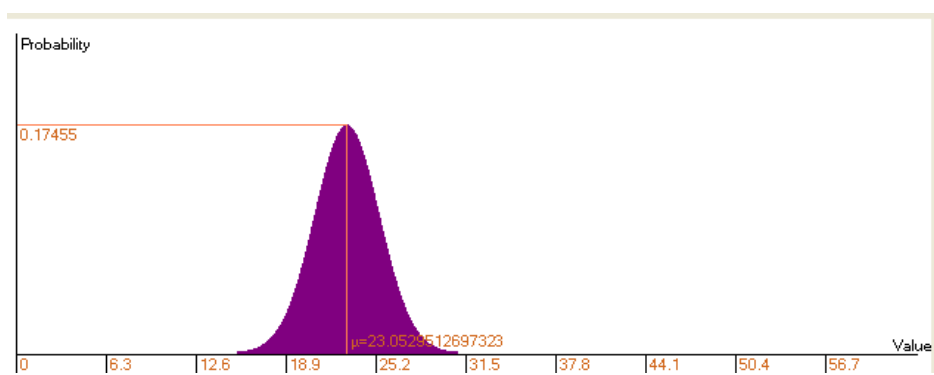

Figure 2-2 Predicted distribution of employed population in 2009

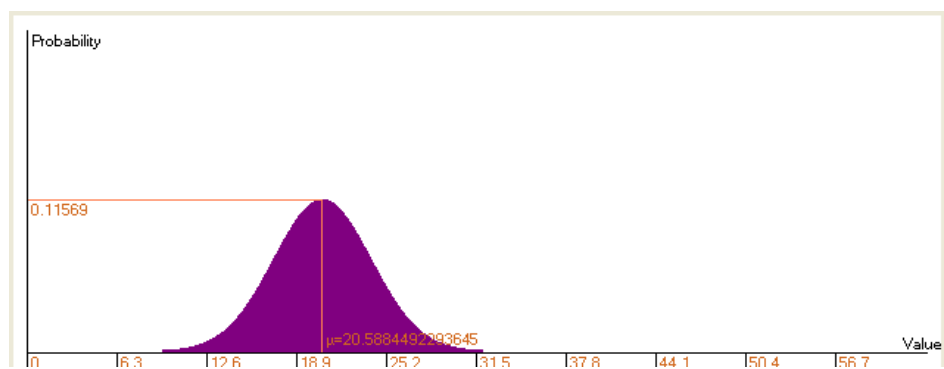

Figure 2-3 Predicted distribution of employed population in 2011

Figure 2-1 and figure 2-2 shows the predicted Gaussian distribution of employed population value for year 2009 and 2011 respectively.

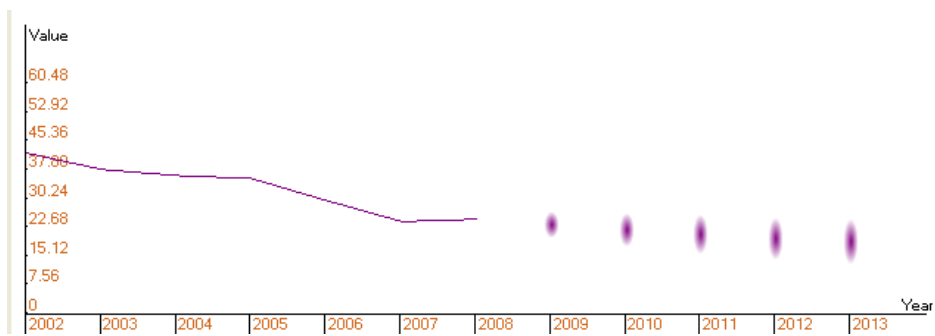

Figure 2-4 History and predicted trend graph for manufacturing employed population from 2002 to 2013

In figure 2-3, history data points are connected with solid line, while predicted distributions are shown as gradient bars: the central point is the expectation value and the bar covers the value range [μ-3σ, μ+3σ].

Another demonstration is on gross revenue of gambling industry based on history data from 2002 to 2009:

| Year | 2002 | 2003 | 2004 | 2005 |
|------|------|------|------|------|
| **Gross Revenue of Gambling (million MOP)** | 23496.0 | 30315.1 | 43510.9 | 47133.7 |
| **Year** | 2006 | 2007 | 2008 | 2009 |
| **Gross Revenue of Gambling (million MOP)** | 57521.3 | 83846.8 | 109826.3 | 120383.0 |

Table 2-2 History data of gambling gross revenue

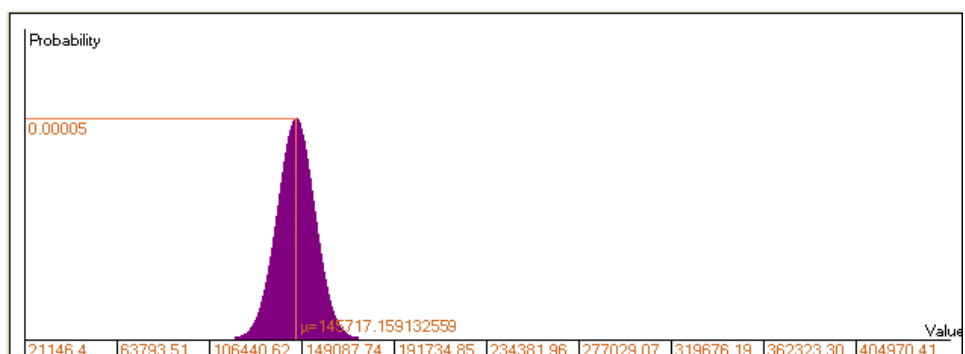

Figure 2-5 Predicted distribution of gambling gross revenue in 2009

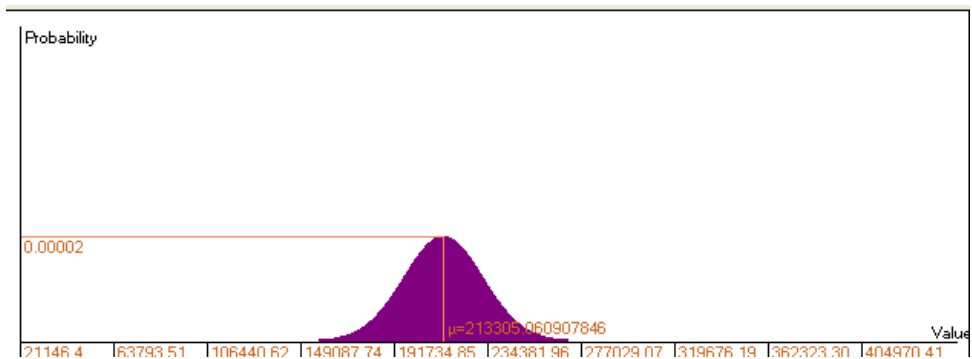

Figure 2-6 Predicted distribution of gambling gross revenue in 2012

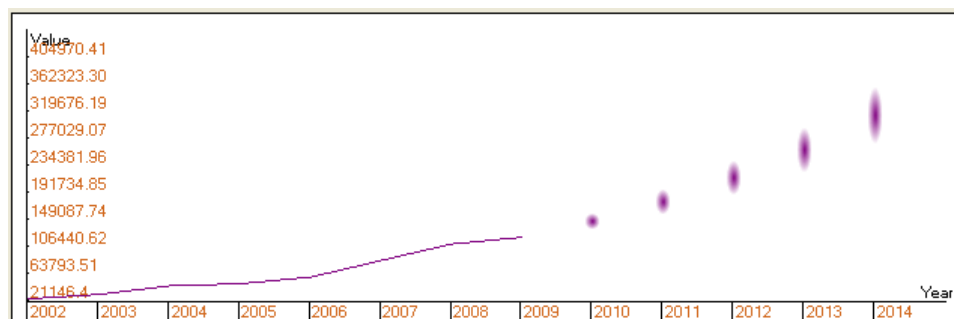

Figure 2-7 History and predicted trend graph for gambling gross revenue from 2002 to 2014

By using prediction for continuous variable, discretizing of the continuous indices could be avoided and more specific results than level value will be available. However, the prediction strongly depends on the distribution and function model we choose.

# 3   Correlation Analysis among Indices

In the discussion of dynamic decision process in chapter 1, it is supposed that the action node is readily available: the problem has already defined several candidate actions and the decision-maker only has to choose one for each state encountered. However, sometimes the decision-maker has to figure out what are the feasible actions first rather than just choose.

This is exactly the case in Macau industry diversification problem where no readily-made action collections are available. What's more, the number of possible actions is so huge that we probably can never traverse the solution space to test each solution and find the best policy. Besides, the industry diversification is a complicate economic problem which can hardly be handled without human experts' interference.

Thereafter, as the foundation of the whole dynamic decision process, firstly a mechanism should be provided to help decision-makers to define the action collection. This paper considers using correlation analysis as such a tool.

Correlation analysis among multiple indices reveals relationship information, such as which indices are closely related, which index influences the index of interest the most, how will changing of some indices affect others, etc. It could assist decision-makers to realize relationship among factors more clearly. Most importantly, decision-makers can use this information to adjust indices which are relatively easy to control and finally achieve regulation of certain indices of interest.

There are several ways to do correlation analysis. Linear or non-linear correlation could be measured, and non-linear correlation can be further divided into monotonic and non-monotonic correlation.

## 3.1 Pearson Linear Correlation Coefficient

One of the most widely used correlation analysis method is Pearson correlation coefficient [5]. It represents the linear correlation between two random variables, and could be easily extended to represent linear correlation of multiple variables and partial correlation of two variables with elimination of effects from other relative variables.

The formula of Pearson correlation coefficient for two random variables is

$$r = \frac{\sum_{i=1}^{n}(x_i - \bar{x})(y_i - \bar{y})}{n\sigma_x\sigma_y} \qquad (3.1)$$

It is clear that two linear correlated variables have the following properties: 1) the intensity of the data points around the regression line; 2) the correspondence direction of correlation, which is, whether the two variables change in accordance or in contrary.

From the formula it could be seen that $(x_i - \bar{x})(y_i - \bar{y})$ reflects both the degree and direction of correlation: 1) the larger the product is, the bigger the degree of correlation is; 2) By using $(\bar{x}\bar{y})$ as the origin, positive/negative products indicate positive/negative correlation respectively. The "divide by product of standard deviation" operation can eliminate effect of different dispersion from different statistical data and make the coefficient comparable. The computed coefficient is a number in [-1, +1] and larger absolute value stands for stronger linear correlation.

For multiple variables, after calculating correlation coefficient between each pair of variables, multivariant correlation coefficient can be calculated between the dependent variable and multiple independent variables and then compute partial correlation between the dependent variable and each of the independent variables, which allows us to select independent variables that have significant effect on the dependent variable step by step.

Despite of its convenience, the application of Pearson correlation coefficient is strictly limited to linear correlation. Though two independent variables should have a zero correlation coefficient, a zero correlation coefficient does not necessarily mean that the variables are independent.

## 3.2  Monotonic Correlation

Considering the function $y = x^2$, $x > 0$. It is obvious that x and y are totally correlated, but if we compute linear coefficient on a random sampling of points on the function curve, the result could be much smaller than what we expect.

Monotonic correlation measures the extent to which that two variables change correspondingly without requiring a linear change rate. This is also called rank correlation. Two commonly used rank correlation coefficients will be described in this paper: Spearman's rank correlation coefficient and Kendall's tau rank correlation coefficient [6]. Both of these coefficients lie in [-1, +1], with +1 for perfect ranking agreement, -1 for perfect ranking disagreement and 0 for perfect ranking independence.

The Spearman correlation coefficient looks like the ranked variable version of Pearson correlation coefficient. In order to compute this coefficient, the variable values are replaced by their rank values.

For example, the history (x, y) values are A: (0, 2), B: (3, 7) and C: (-5,-1). The first step is to reorder the points according to x and y respectively, getting $x_B > x_A > x_C$ and $y_B > y_A > y_C$. Then the original data is replaced by: A: (2, 2), B: (1, 1) and C: (3, 3), which are the rank values. Finally, the Pearson correlation coefficient formula (3.1) is used on rank values to compute the result. In fact, if there are no replicate values in x and y, there is another simpler formula

$$\rho = 1 - \frac{6\sum d_i^2}{n(n^2-1)} \qquad (3.2)$$

where $d_i$ is the difference between each pair of rank values $(x_i, y_i)$.

Kendall's Tau coefficient is named after its inventor Maurice Kendall, defined as

$$\tau = \frac{n_c - n_d}{\frac{1}{2}n(n-1)} \qquad (3.3)$$

where $n_c$ is the number of concordant pairs (sign of $x_i$ - $x_j$ is equal to sign of $y_i$ - $y_j$), and $n_d$ is the number of discordant pairs (sign of $x_i$ - $x_j$ is different from sign of $y_i$ - $y_j$) in the data set. A bigger absolute value of $n_c - n_d$ means that concordant/discordant pairs have a larger proportion in all the pairs of items and rankings are more consistent.

Rank correlation analysis is not applicable for all non-linear correlations. A simple example is $y = x^2$, $x \sim N(0,1)$. Both linear correlation coefficient and rank correlation coefficient will yield an average value of zero.

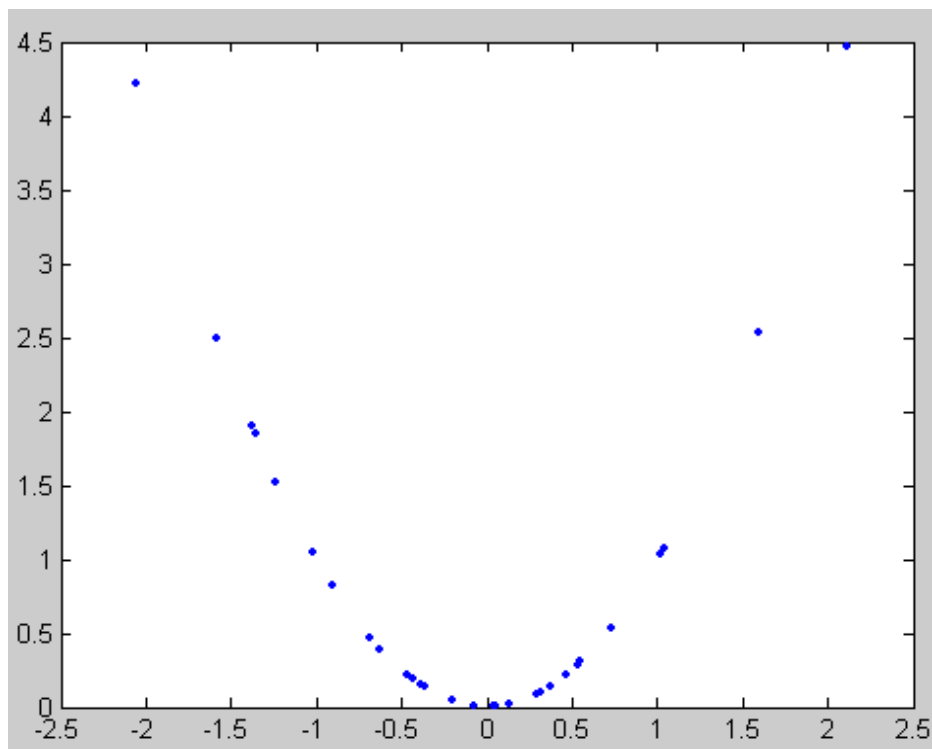

Figure 3-1 Thirty sample points with $y = x^2$, $x \sim N(0,1)$

This is because non-linear correlations do not always have a determinate correlating direction. From figure 3-1 it can be seen there is a inflection point at (0, 0) for quadratic curve $y = x^2$, and the ranking consistency is exactly opposite on each side of the point. Moreover, since there are a great variety of curve families, analysts usually cannot subjective pick one kind of curve as standard to analyze the correlation.

## 3.3  General Non-monotonic Correlation

In the general case, we should measure the correlation degree of data without considering the direction of correlation.

One method is correlation ratio [7][8]. It groups the data according to one of the variables, measuring the dispersion inside each group and among different groups to reveal the intensity of data points around the distribution curve. For instance, the data pairs (x, y) are divided into k groups according to value of x. We use $y_{i1}$, $y_{i2}, \dots y_{in_i}$ to denote y values in group i and $n_i$ to denote the number of data points group i has. After computing mean value of all y values and of each group, the correlation ratio of y to x is

$$e_{yx}^2 = \frac{\sum_i n_i (\overline{Y}_i - \overline{Y})^2}{\sum_i \sum_j (Y_{ij} - \overline{Y})^2} \qquad (3.4)$$

Correlation ratio also has a limited scope [7]. Moreover, if the data collection is small, it is rather difficult to obtain enough groups.

There are also analysis methods based on information entropy, like total correlation:

$$C(X_1, X_2, \dots, X_n) = \sum_{i=1}^{n} H(X_i) - H(X_1, X_2, \dots, X_n) \qquad (3.5)$$

where $H(X_i)$ is the information entropy of variable $X_i$ and $H(X_1, X_2, \dots, X_n)$ is the joint entropy of all the variables.

For discrete random variables, the entropy is relatively easy to compute. However, for continuous variables, the probability density function must be specified in order to compute its entropy, which, is not easy to determine in the application proposed in this paper. Though discretizing of the continuous variables could be applied, a small history data would seriously affect the reliability of the result.

## 3.4  Demonstration

Although the scope of linear correlation coefficient and monotonic correlation coefficients are quite limited, considering the complexity and subjectivity of general non-linear correlation analysis, the correlation among industrial indices in a relatively short time span can be approximately treated as linear or monotonic.

The implementation provided in this paper uses Pearson correlation coefficient, Kendall's tau correlation coefficient and Spearman rank correlation to analyze correlation between two indices. Below is the interface of the implementation:

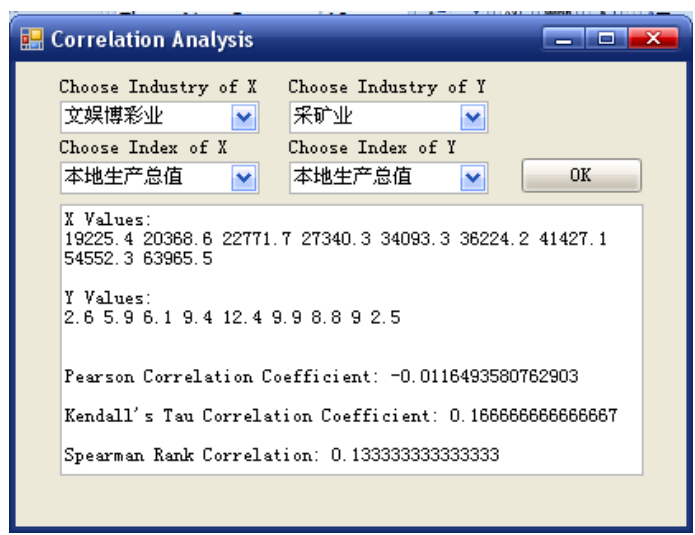

Figure 3-2 Correlation analysis interface

Industries and indices of X and Y are selected from the combo-boxed, and analysis results are shown in the textbox on the lower part of the window.

The following is a demonstration of analyzing the correlation between Macau employed population and tax revenue of gambling industry using seasonal history data from 2004 to 2005:

| Year | 2004 | | | | 2005 | | | |
|---|---|---|---|---|---|---|---|---|
| Season | 1 | 2 | 3 | 4 | 1 | 2 | 3 | 4 |
| Employed Population (thousand) | 20.1 | 22.0 | 23.0 | 26.7 | 27.5 | 28.7 | 33.3 | 33.7 |
| Tax Revenue of Gambling (million MOP) | 3202.0 | 3578.4 | 4232.7 | 4223.5 | 3993.3 | 4524.9 | 4553.4 | 4246.9 |

Table 3-1 History data of employed population and tax revenue of gambling

```
X Values:
20.1 22 23 26.7 27.5 28.7 33.3 33.7

Y Values:
3202 3578.4 4232.7 4223.5 3993.3 4524.9 4553.4 4246.9

Pearson Correlation Coefficient: 0.776017710959035

Kendall's Tau Correlation Coefficient: 0.642857142857143

Spearman Rank Correlation: 0.833333333333333
```

Figure 3-3 Correlation between employed population and tax revenue of gambling

From the result of the program, it can be seen that the results are highly consistent: all three correlation analysis methods give a relatively high positive coefficient, which means the two indices are positively correlated.

Another demonstration is between annual gross domestic product (GDP) of gambling and mining industry from 2000 to 2008:

| Year | 2000 | 2001 | 2002 | 2003 | 2004 | 2005 | 2006 | 2007 | 2008 |
|---|---|---|---|---|---|---|---|---|---|
| Gambling GDP (million MOP) | 19225.4 | 20368.6 | 22771.7 | 27340.3 | 34093.3 | 36224.2 | 41427.1 | 54552.3 | 63965.5 |
| Mining GDP (million MOP) | 2.6 | 5.9 | 6.1 | 9.4 | 12.4 | 9.9 | 8.8 | 9 | 2.5 |

Table 3-2 History data of GDP of gambling and mining

```
X Values:
19225.4 20368.6 22771.7 27340.3 34093.3 36224.2 41427.1
54552.3 63965.5

Y Values:
2.6 5.9 6.1 9.4 12.4 9.9 8.8 9 2.5

Pearson Correlation Coefficient: -0.0116493580762903

Kendall's Tau Correlation Coefficient: 0.166666666666667

Spearman Rank Correlation: 0.133333333333333
```

Figure 3-4 Correlation between GDP of gambling and mining

Since the correlation coefficients computed are pretty close to zero, it can be concluded that the correlation between GDP of gambling and mining is low. Again, the three coefficients are rather consistent.

# 4 Extension and Improvement in DCMS

## 4.1 Appending History Database

In order to append dynamic decision module to DCMS, the database, which only stored participants and templates' information, should be extended to include history data of industry indices.

To Store the history data, three new tables are added to the original database with the following schema (designed by my teammate Miss Wang Juewen):

| Field Name | Data Type | Primary Key | Null Value Allowed |
|---|---|---|---|
| Industry_Id | int | Yes | No |
| Industry_Name | varchar(50) | No | No |
| Industry_Remark | varchar(200) | No | Yes |
| Industry_Type | varchar(50) | No | Yes |
| Industry_Enabled | int | No | No |

Table 4-1 Schema of table History_IndustryID

| Field Name | Data Type | Primary Key | Null Value Allowed |
|---|---|---|---|
| Index_Id | int | Yes | No |
| Index_Name | varchar(50) | No | No |
| Index_Remark | varchar(200) | No | Yes |
| Index_Type | varchar(50) | No | Yes |
| Index_Enabled | int | No | No |

Table 4-2 Schema of table History_IndexID

| Field Name | Data Type | Primary Key | Null Value Allowed |
|---|---|---|---|
| Idata_Id | int | Yes | No |

| Index_Id | int | No | No |
|---|---|---|---|
| Industry_Id | int | No | No |
| Idata_Data | float | No | No |
| Idata_Year | int | No | No |

Table 4-3 Schema of table History_IndustryIndexData

In practice, the history data of Macau industry is downloaded from official website of Macau Statistics and Census Service. However, the data offered on the website is stored in excel spreadsheets, which must be converted into SQL server data tables. Moreover, the format of the data spreadsheets is varied and multiple indices may be stored in one spreadsheet.

In order to perform conversion, a converting tool is developed which writes the content of a spreadsheet into the SQL server database according to a rule file provided by user. The rule file contains several rules, each of them denotes one index stored in the spreadsheet. A rule consists of: industry ID of the index, index ID of the index as well as at which column it is stored in the spreadsheet.

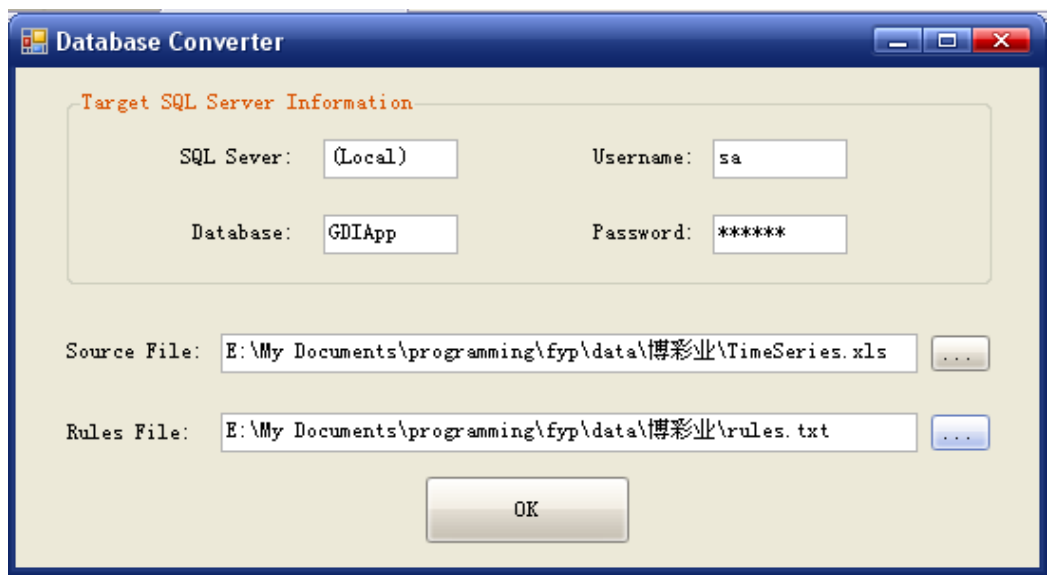

Figure 4-1 Interface of database converter

Shown below are parts of the converted database:

| Industry_Id | Industry_Name | Industry_Remark | Industry_Type | Industry_Ena |
|---|---|---|---|---|
| 1 | 文娱博彩业 | 文娱博彩业 | 第三产业 | 1 |
| 3 | 酒店及饮食业 | 酒店及饮食业 | 第三产业 | 1 |
| 5 | 金融保险业 | 金融保险业 | 第三产业 | 1 |
| 6 | 制造业 | 制造业 | 第二产业 | 1 |
| 7 | 建筑业 | 建筑业 | 第二产业 | 1 |
| 8 | 水电及气体生… | 水电及气体生… | 第二产业 | 1 |
| 9 | 批发及零售业 | 批发及零售业 | 第三产业 | 1 |

Figure 4-2 Table History_IndustryID

| Index_Id | Index_Name | Index_Remark | Index_Unit | Index_Enable |
|---|---|---|---|---|
| 1 | 本地生产总值 | 按支出法估算 | 百万澳门元 | 1 |
| 2 | 税总收入 | | 百万澳门元 | 1 |
| 3 | 就业人口 | 按行业划分 | 千人 | 1 |
| 4 | 毛收入 | | 百万澳门元 | 1 |
| 5 | 每月工作收入 | 按行业划分 | 澳门元 | 1 |
| 6 | 本地生产总值 | 按生产方法估算 | 百万澳门元 | 1 |
| 9 | 就业人口同期… | 按行业划分 | % | 1 |
| 10 | 就业人口同期… | 按行业划分 | 暂无 | 1 |
| 11 | 人文发展指数点 | 人文发展指数点 | | 1 |
| 12 | 水资源消耗 | 水资源消耗 | 千立方米 | 1 |
| 13 | 电力消耗 | 电力消耗 | 百万千瓦小时 | 1 |
| 14 | 石油气消耗 | 石油气消耗 | 公吨 | 1 |

Figure 4-3 Table History_IndexID

| Idata_Id | Index_Id | Industry_Id | Idata_Data | Idata_Year |
|---|---|---|---|---|
| 57 | 6 | 5 | 9545.1 | 2002 |
| 58 | 6 | 5 | 10205.2 | 2003 |
| 59 | 6 | 5 | 11663.6 | 2004 |
| 60 | 6 | 5 | 16123.6 | 2005 |
| 61 | 6 | 5 | 21276.3 | 2006 |
| 62 | 6 | 5 | 26516.5 | 2007 |
| 63 | 6 | 5 | 29448 | 2008 |
| 64 | 6 | 1 | 19225.4 | 2000 |
| 65 | 6 | 1 | 20368.6 | 2001 |
| 66 | 6 | 1 | 22771.7 | 2002 |
| 67 | 6 | 1 | 27340.3 | 2003 |
| 68 | 6 | 1 | 34093.3 | 2004 |
| 69 | 6 | 1 | 36224.2 | 2005 |
| 70 | 6 | 1 | 41427.1 | 2006 |
| 71 | 6 | 1 | 54552.3 | 2007 |
| 72 | 6 | 1 | 63965.5 | 2008 |
| 73 | 12 | 0 | 48846 | 2000 |
| 74 | 12 | 0 | 48374 | 2001 |
| 75 | 12 | 0 | 49078 | 2002 |
| 76 | 12 | 0 | 51628 | 2003 |
| 77 | 12 | 0 | 53392 | 2004 |
| 78 | 12 | 0 | 55860 | 2005 |
| 79 | 12 | 0 | 60357 | 2006 |

Figure 4-4 Table IndustryIndex_Data

## 4.2  Rearrangement of DCMS Graphic Classes

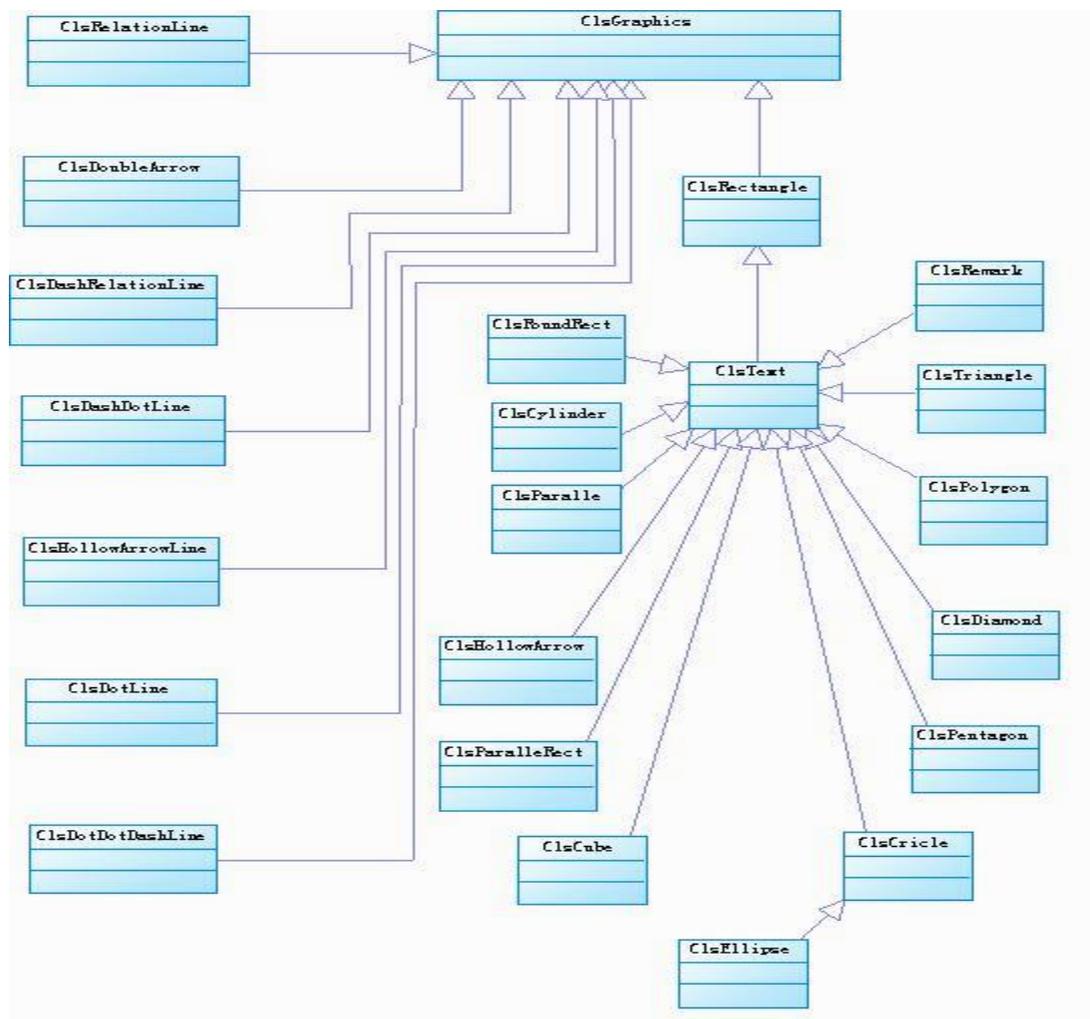

Figure 4-5 Original graphic class diagram of DCMS

Several problems could be found from the original graphic class diagram of DCMS shown in figure 4-1. There are seven classes representing line that directly inherit from ClsGraphics, which is the base class for all graphic elements. However, immediate perception tells us these line classes probably have lots of attributes and functions in common, which are not handled in the base class, and by grouping these classes under a common Line super class should lead to a more natural designing and provide convenience in modification. A further look into these classes reveals almost the same contents, except for a few displaying details.

Therefore, a class ClsLine is added as a subclass of ClsGraphics and super class for all the seven line classes, and a function for set displaying parameters is included in ClsLine to be overloaded in subclasses to implement drawing detail respectively.

Below is the rearranged class diagram:

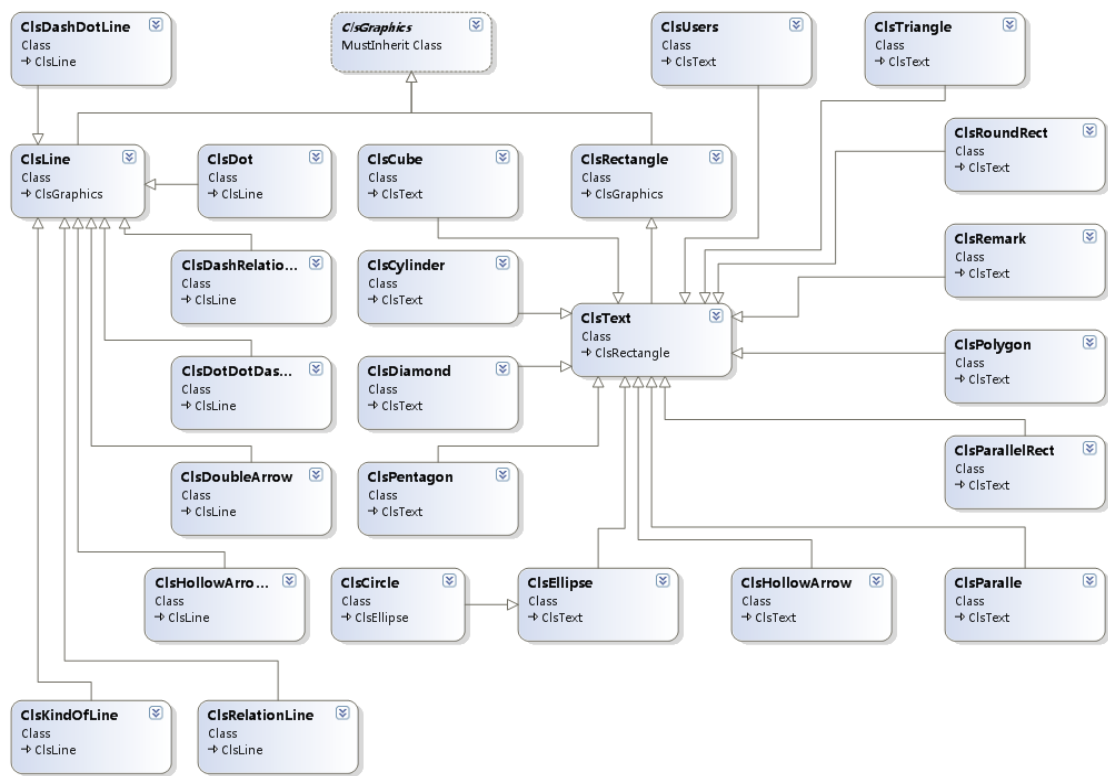

Figure 4-6 Rearranged graphic class diagram of DCMS

Since each of the subclasses is still an instance of ClsGraphics, this rearrangement
hardly needs modification at other part of the system. Still, the necessity of so
many different subclasses for line is doubtable. If the difference between a
subclass and the base class is simply line style, an attribute in the base class can
easily fulfill the task. Nevertheless, as a modeling system, these different lines
may represent different semantic meanings, which, though not have according
implementation now, might serve certain functionalities in future updates.

## 4.3　Curve Implementation

Curves make the template with lots of relation lines more clear. Cubic Bezier
curve is used for implementation, which is well support in GDI+ in visual
studio .NET.

The parameter function for cubic Bezier curve is:

$$B(t) = (1-t)^3 P_0 + 3(1-t)^2 t P_1 + 3(1-t)t^2 P_2 + t^3 P_3, t \in [0,1] \quad (4.1)$$

where $P_0$ and $P_3$ are start and end point respectively, while $P_1$ and $P_2$ are anchor points providing direction information.

The simplest way to generate such a curve is to loop a series of t values through the zone [0, 1], compute according X(t) and Y(t) and connect computed points. Hit-testing can be done in a similar way: just use less number of points to generate a flattened curve, compute the distance from the given hit point to each of the straight lines composing the flattened curve and test the minimum distance. A considerable defect of such an approach is that while points may be spaced too far apart where the curvature is large, too many points may be generated in areas where the curve is close to linear. An alternative approach is recursive subdivision which makes use of Bezier curve's property that a curve can be split at any point into 2 subcurves, each of which is also a Bezier curve.

The good news is that GDI+ provides direct support for Bezier curves so it is not needed to build up the curve from scratch in the actual implementation. The related GDI+ members include:

- Graphics.DrawBezier Method (Pen, Point, Point, Point, Point): Draws a Bézier spline defined by four Point structures.
- System.Drawing.Drawing2D.GraphicsPath Class: Represents a series of connected lines and curves. In our implementation this class and its members serve hit-testing.
- GraphicsPath.AddBezier Method (Point, Point, Point, Point): Adds a cubic Bézier curve to the current figure.

Below is several screenshots demonstrating the implemented curve relation-line. After selecting and right-click a straight line:

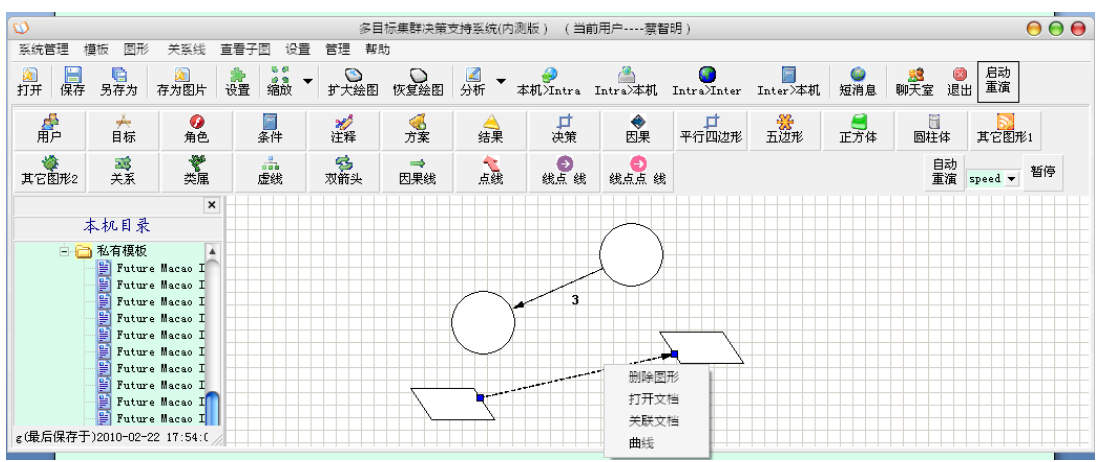

Figure 4-7 Context Menu for Straight Line

After checking the "Curve" menu item, two anchor points appeared:

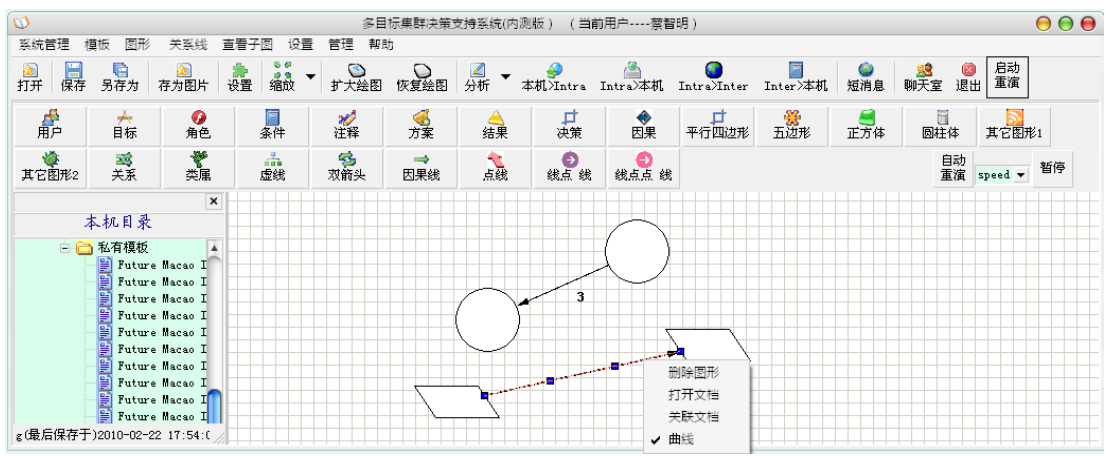

Figure 4-8 Context Menu for Curve

After dragging the anchor points, the shapes of curves change accordingly:

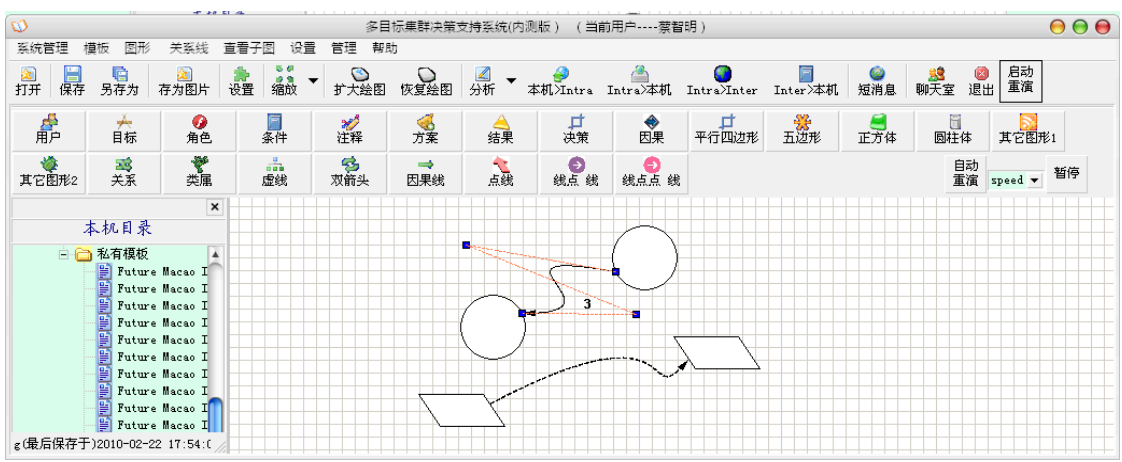

Figure 4-9 Modified Curves

# 5   Conclusions and Future Work

This thesis discusses dynamic decision process extension for DCMS along with its application in Macau industry diversification problem. Two popular methods for modeling dynamic decision process, Markov decision process and dynamic Bayesian decision network are introduced. Implemented prediction in Markov process and correlation analysis among indices for decision making are discussed in detail, as well as some improvement in DCMS.

MDP represents the whole system as a single state variable, where the next state is only related to the current state and action taken. Prediction for both discrete and random state variables is discussed. With the action node determined, it will be convenient to compute optimal policy in MDP using value iteration.

Correlation analysis yields the relationship among indices and help people to design the decision. In practice, correlation analysis of industry factors on a relatively short time span could be considered as linear or monotonic correlation. Pearson correlation efficient, Spearman rank correlation coefficient as well as Kendal's tau rank correlation coefficient between two random variables are implemented.

By rearranging graphic class hierarchy and implementing curve relation-line, the class relationship of the system is made clearer and further strengthened the functionality of the modeling system.

Extending the DCMS with dynamic decision module requires a lot of modification and integration. A first step is adding history data of Macau industry into the original database. The next step is to adjust the original template modeling function in DCMS to represent the dynamic decision process as well as convenient action/reward node generating mechanism from user input data, which requires appropriate design of database as well as graphical element semantic.

Mining correlation may also help discover causal relationships among indices, which could in turn help design a Bayesian network structure for better representing of factor relationships.

# Index of Figures

# Index of Tables

# Acknowledgement

I would like to acknowledge all those people who have generously offer help and support to my final year project.

First of all, I would like to express my sincere gratitude and special thanks to my supervisor, Prof. Cai Zhiming. He was always patient and helpful during my work, and he provided me a lot of instructive advises. His encouragement and trust really helped me to finish my study successfully.

Also, I would like to thank Mr. Huang Liangli and Mr. Gu Yinshen, who has provided a lot of practical and helpful advices for my study. Furthermore, I would like to thank my teammates in the project for their constant support, especially Miss Wang Juewen, who has defined the schema of history database which is the foundation of my database conversion work in section 4.1.

Finally, I would like to thank Macau University of Science and Technology (MUST) for offering me a good environment to study.